%% file: main.tex
\pdfoutput=1

\documentclass[11pt]{article}

\usepackage[preprint]{acl}

\usepackage{times}
\usepackage{latexsym}

\usepackage[T1]{fontenc}

\usepackage[utf8]{inputenc}

\usepackage{microtype}

\usepackage{inconsolata}

\input{pkg.tex}

\title{Holistic Audit Dataset Generation for LLM Unlearning via Knowledge Graph Traversal and Redundancy Removal}

\author{
\textbf{Weipeng Jiang}\textsuperscript{1} ,
\textbf{Juan Zhai}\textsuperscript{2},
\textbf{Shiqing Ma}\textsuperscript{2}, 
\textbf{Ziyan Lei}\textsuperscript{1}, \\
\textbf{Xiaofei Xie}\textsuperscript{3},
\textbf{Yige Wang}\textsuperscript{1},
\textbf{Chao Shen}\textsuperscript{1} 
\\
\normalsize{
\textsuperscript{1}Xi'an Jiaotong University,
\textsuperscript{2}University of Massachusetts Amherst
\textsuperscript{3}Singapore Management University}
\\
\normalsize{\texttt{\{lenijwp@stu, l13201738997@stu, jihejue039@stu, chaoshen@mail\}.xjtu.edu.cn}}
\\
 \normalsize{\texttt{\{juanzhai, shiqingma\}@umass.edu}}
\\
\normalsize{\texttt{xfxie@smu.edu.sg}}
}

\begin{document}
\maketitle

\input{src/abstract}
\input{src/intro}
\input{src/preli}

\input{src/method}

\input{src/exp}
\input{src/rw}
\input{src/conclusion}
\input{src/limitation}


\bibliography{ref}

\appendix

\input{src/appendix}



\end{document}

%% file: pkg.tex
\usepackage[utf8]{inputenc} 
\usepackage[T1]{fontenc}    
\usepackage{booktabs}       
\usepackage{amsfonts}       
\usepackage{nicefrac}       
\usepackage{microtype}      
\input{todo.tex}

\usepackage{xspace}
\usepackage{multirow}
\usepackage{graphicx}
\usepackage{algorithm}
\usepackage{amsmath, amsthm, amssymb}
\usepackage{algorithmicx}
\usepackage[noend]{algpseudocode}
\usepackage{bm}
\usepackage[most]{tcolorbox}

\usepackage{longtable}
\usepackage{subfigure}
\usepackage{tabularx} 

\usepackage{cleveref}
\newcommand{\sys}{\mbox{\textsc{HANKER}}\xspace}

\usepackage{pifont} 
\usepackage{array}
\newtcolorbox{colorquote}[1][]{
    boxrule=0.5pt,
    left=1pt,
    right=1pt,
    top=1pt,
    bottom=1pt,
    colback=black!5,
    colframe=black!55,
    notitle,
    enhanced,
    breakable,
}
\definecolor{myred}{RGB}{224,0,0}
\definecolor{myblue}{RGB}{46,117,182}
\definecolor{mygreen}{RGB}{83,130,53}
\definecolor{myyellow}{RGB}{191,144,0}
\usepackage{mdframed}
\newmdenv[linewidth=1pt, linecolor=blue, backgroundcolor=gray!20, roundcorner=10pt]{myframe}
\usepackage{cuted}             
\usepackage{listings}

%% file: todo.tex
\newif\ifrevise
\newif\ifrevisenew
\newif\iffinal

\revisetrue
\revisenewtrue
\finaltrue

\revisefalse
\revisenewfalse
\finalfalse

%% file: src/abstract.tex
\begin{abstract}

    In recent years, Large Language Models (LLMs) have faced increasing demands to selectively remove sensitive information, protect privacy, and comply with copyright regulations through unlearning, by the \textit{Machine Unlearning}. While evaluating unlearning effectiveness is crucial, existing benchmarks are limited in scale and comprehensiveness, typically containing only a few hundred test cases. 
    We identify two critical challenges in generating holistic audit datasets: ensuring audit adequacy and handling knowledge redundancy between forget and retain dataset. To address these challenges, we propose \sys, an automated framework for holistic audit dataset generation leveraging knowledge graphs to achieve fine-grained coverage and eliminate redundant knowledge. Applying \sys to the popular MUSE benchmark, we successfully generated over 69,000 and 111,000 audit cases for the News and Books datasets respectively, identifying thousands of knowledge memorization instances that the previous benchmark failed to detect. 
    Our empirical analysis uncovers how knowledge redundancy significantly skews unlearning effectiveness metrics, with redundant instances artificially inflating the observed memorization measurements ROUGE from 19.7\% to 26.1\% and Entailment Scores from 32.4\% to 35.2\%, highlighting the necessity of systematic deduplication for accurate assessment.

\end{abstract}

%% file: src/intro.tex
\section{Introduction}
\label{sec:intro}

In recent years, Large Language Models (LLMs) have undergone rapid development, demonstrating impressive capabilities across a wide range of applications, from natural language processing to code generation and complex problem-solving~\cite{liu2023codegeneratedchatgptreally, satpute2024can}.
However, these advances have raised concerns about potential risks associated with the vast knowledge stored in these models, e.g., the inadvertent retention of personally identifiable information (PII)~\cite{jang2022knowledge}, the propagation of unsafe or biased behaviors~\cite{liu2024saferlargelanguagemodels}, and the unauthorized use of copyrighted content~\cite{eldan2023s}.
Furthermore, there is an increasing imperative for LLMs to comply with regulatory standards such as the General Data Protection Regulation (GDPR)~\cite{hoofnagle2019european}, which enforces the ``Right to be Forgotten''~\cite{dang2021right}.
To address these concerns, researchers are investigating various unlearning techniques~\cite{jia2024soul} to selectively remove specific knowledge from pre-trained LLMs while preserving their general language modeling capabilities, thereby avoiding the substantial computational costs associated with building new models from scratch.

\input{tfsrc/fig_musecase.tex}

\input{tfsrc/fig_overalltask.tex}

The growing significance of LLM unlearning has heightened the importance of rigorous evaluation or audit of unlearing performance. Recent benchmarks like MUSE~\cite{shi2024muse} and TOFO~\cite{maini2024tofu} assess unlearning efficacy across multiple dimensions, ranging from verbatim text retention to embedded knowledge preservation. 
These pioneering frameworks have advanced the field by establishing standardized datasets, providing pre-trained target models, and introducing multifaceted evaluation metrics. However, their audit suites remain constrained in scope—for instance, MUSE employs only 100 test questions to evaluate 0.8M corpora. From an auditing perspective, such limited test coverage may inadequately assess the targeted knowledge removal, potentially compromising the comprehensive evaluation of unlearning effectiveness.

Our investigation reveals two fundamental challenges in holistic audit dataset synthesis. The primary concern about \textit{\textbf{audit adequacy}} stems from  simply relying on GPT-4 for automated QA generation from forget corpora. While this approach can generate multiple question-answer pairs for each target text, it introduces significant uncertainty in whether the generated questions comprehensively cover all the critical information contained within the source text.
The second challenge involves \textbf{\textit{knowledge redundancy}} between forget and retain corpora. As illustrated in \autoref{fig:overalltask}, shared knowledge should be preserved during the unlearning process. However, current evaluation methods fail to account for test cases where the information targeted also appears in the retain dataset, as demonstrated in \autoref{fig:musecase}.

In this paper, we propose \sys, a novel automated framework for holistic audit dataset generation that leverages knowledge graphs (KGs) to address the aforementioned limitations.
Benefiting from advances in named entity recognition and information extraction, various tools now enable efficient conversion of unstructured text into structured entity-relation graphs. 
\sys first converts both forget and retain corpora into structural knowledge graphs. By treating each KG edge (i.e., one fact) as a minimal unit, we can explicitly control the coverage of the audit process. 
Subsequently, by identifying and eliminating identical facts within the forget and retain KGs, we remove redundant knowledge from the forget KG, ensuring a well-defined audit scope. 
Finally, \sys utilizes specific facts to guide LLMs in generating high-quality, targeted test questions, guaranteeing comprehensive and accurate auditing.
Through this pipeline, \sys automatically generates large-scale, comprehensive audit datasets for any given forget and retain corpora, thereby providing robust support for LLM unlearning evaluation.

In summary, our contributions are as follows:
\begin{itemize}
    \item We introduce \sys\footnote{\url{https://anonymous.4open.science/r/HANKER-FB86}}, a novel and automated framework for generating holistic audit datasets for LLM knowledge unlearning, which addresses the challenge of \textit{audit adequacy} and \textit{knowledge redundancy}.
    \item We apply \sys to popular benchmark MUSE, significantly expanding the dataset scale and identifying knowledge memorization cases in unlearned LLMs that exceeded previous findings by three orders of magnitude ($10^3\times$).
    \item Our experimental results reveal that knowledge redundancy has a substantial impact on the assessment of unlearning effectiveness.
\end{itemize}

%% file: tfsrc/fig_musecase.tex
\begin{figure}[!tbp]
    \centering     
    \includegraphics[width=\linewidth]{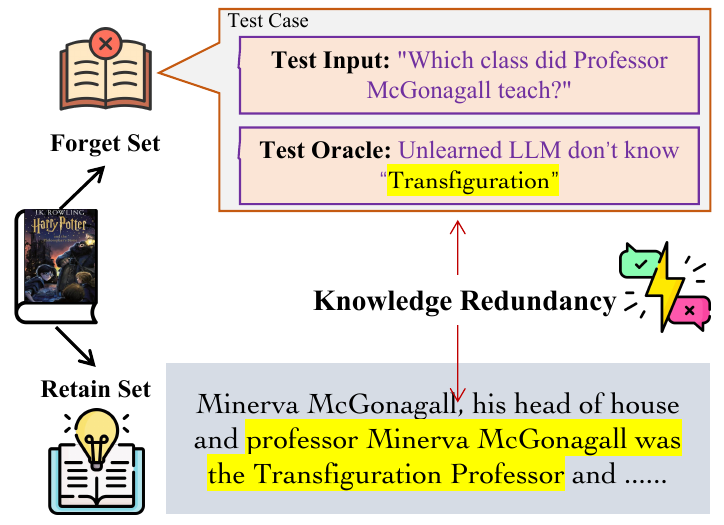}
    \caption{An illustrative example from MUSE demonstrating where knowledge targeted for forgetting also appears in the Retain Dataset, highlighting the challenge of knowledge redundancy in unlearning evaluation.}
    \label{fig:musecase}
\end{figure}

%% file: tfsrc/fig_overalltask.tex
\begin{figure*}[!t]
    \centering     
    \includegraphics[width=0.85\linewidth]{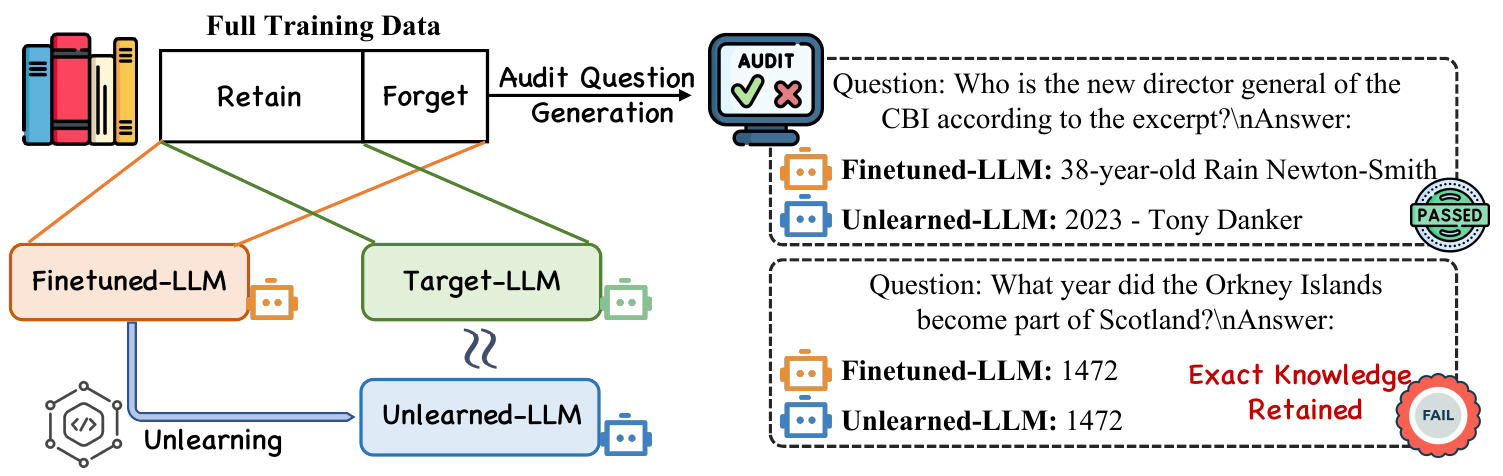}
    \caption{Illustration of the basic pipeline for LLM knowledge unlearning and its audit. 
    }
    \label{fig:overalltask}
\end{figure*}

%% file: src/preli.tex
\section{Preliminaries and Motivation}
\label{sec:prelim}

\subsection{LLM Unlearning}

LLM unlearning refers to techniques that selectively remove specific behaviors or knowledge from a pre-trained language model while maintaining its overall functionality~\cite{yao2023large}. 
With the proliferation of LLMs, unlearning has gained significant attention due to its broad applications in safety alignment, privacy protection, and copyright compliance~\cite{eldan2023s,liu2024rethinking,jia-etal-2024-soul}. The evaluation and auditing of LLM unlearning spans from basic verbatim memorization to deeper knowledge memorization~\cite{shi2024muse}, with this work focusing on the latter.
As depicted in \autoref{fig:overalltask}, LLM unlearning operates as a targeted intervention within the model's knowledge representation framework. 
Its core objective is the selective removal of specific information while preserving the model's broader knowledge base (e.g, on retain set). 
This study focuses on the knowledge unlearning auditing that assesses unlearned models' behaviors through comprehensive audit cases. Given access to both forget and retain corpora, we generate a holistic set of test questions with reference answers to thoroughly evaluate whether an unlearned model exhibits any residual knowledge memorization.

\subsection{Knowledge Graph}
\label{sec:pre_kg}

A knowledge graph (KG) is a structured multi-relational graph~\cite{bordes2013translating}, usually representing a collection of facts as a network of entities and the relationships between entities.
Formally, a KG \(\mathcal{G} = \langle \mathcal{E}, \mathcal{R}, \mathcal{F} \rangle\) could be considered a directed edge-labeled graph~\cite{ji2021survey}, which comprises a set \(\mathcal{E}\) of entities (e.g., \textit{Harry Potter}, \textit{Hogwarts School}), a set \(\mathcal{R}\) of relations (e.g., \textit{attends}), and a set $\mathcal{F}$ of facts. A fact is a triple containing the head entity \(e_1 \in \mathcal{E}\), the relation $r \in \mathcal{R}$, and the tail entity \(e_2 \in \mathcal{E}\) to show that there exists the relation from the tail entity to the head entity, denoted as \((e_1, r, e_2) \in \mathcal{F}\)~\cite{hogan2021knowledge}. To illustrate, the fact (\textit{Harry Potter}, \textit{attends}, \textit{Hogwarts School}) shows that there exists the \textit{attends} relation between \textit{Harry Potter} and \textit{Hogwarts School}, which indicates``Harry Potter attends Hogwarts School''.

\input{tfsrc/fig_overview.tex}

\subsection{Motivation}
This section aims to illustrate why and how we consider employing KG to facilitate the holistic LLM unlearning audit. 
Two critical factors underpin this task.
\ding{182}\textbf{Audit Adequacy}: The Forget Dataset is an extensive, unstructured corpus. Existing approaches typically rely on the LLM's prior knowledge to directly generate QA pairs or segment the corpus and feed these segments to ChatGPT for automated QA pair generation. Such methods often fail to intuitively reflect or guarantee the sufficiency of the generate dataset.
\ding{183}\textbf{Knowledge Redundancy}: A more subtle and easily overlooked issue is that the Retain Dataset and Forget Dataset may contain overlapping knowledge. As illustrated in \autoref{fig:overalltask}, this overlapping knowledge should be retained by the unlearned model and, therefore not be treated as candidates for the unlearning efficacy audit. Existing evaluation benchmarks like MUSE often neglect this aspect, as evidenced by \autoref{fig:musecase}.

A KG can offer an effective solution to address these two challenges. 
First, the KG inherently captures the knowledge facts within the Forget Dataset at a fine-grained level, with each edge representing a minimal testable unit. 
By ensuring coverage of every edge in the KG, one can achieve a more intuitive and relatively comprehensive audit. 
Moreover, the structured data provided by the KG can facilitate the identification of identical knowledge facts present in both the Retain and Forget Datasets.
This capability allows for refinement of the initial forget knowledge graph by removing potentially retained information.
Finally, owing to recent advances in KG extraction technology, numerous automated extraction models and pipelines are available to support the automated construction of an audit dataset.

%% file: tfsrc/fig_overview.tex
\begin{figure*}[!tbp]
    \centering     
    \includegraphics[width=0.95\linewidth]{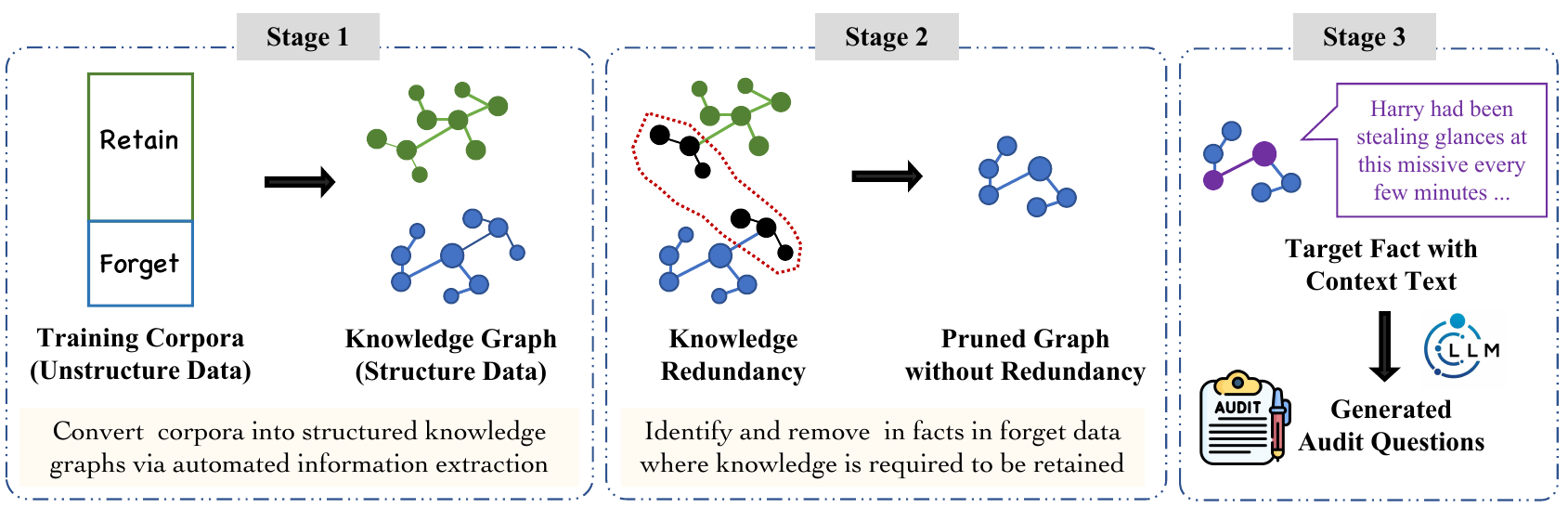}
    \caption{Overview of the proposed \sys. The framework consists of three stages: (1) \textbf{Knowledge Graph Construction} that extracts structured knowledge from forget and retain data, (2) \textbf{Redundancy Removal} that identifies and removes redundant knowledge from the constructed knowledge graphs, and (3) \textbf{Question Synthesis} that generates QA pairs with the guidance of specific facts with LLMs automatically.}
    \label{fig:overview}
\end{figure*}

%% file: src/method.tex
\section{Proposed Method}
\label{sec:method}

The core idea behind \sys is to leverage knowledge graphs to achieve fine-grained and comprehensive test coverage, while rigorously eliminating redundancy between the forgetting and retain objectives. 
As illustrated in \autoref{fig:overview}, \sys comprises three sequential stages. 
During the \textit{Knowledge Graph Construction} stage, unstructured textual data is systematically transformed into structured knowledge representations. This enables the explicit modeling of atomic knowledge units and their semantic interconnections.
Subsequently, the \textit{Redundancy Removal} stage meticulously identifies and eliminates knowledge facts that are simultaneously present in both forget and retain datasets.
This process helps prevent inaccurate assessments by ensuring the audit doesn't mistakenly flag knowledge meant for retain as candidates for removal. 
Finally, in the \textit{Question Synthesis} stage, \sys employs LLMs to generate targeted questions and corresponding reference answers, guided by specific knowledge facts from the pruned knowledge graph. 
This approach provides an automated and holistic evaluation framework for assessing LLM knowledge unlearning efficacy.

\input{tfsrc/algo_method.tex}

\subsection{Stage 1: Knowledge Graph Construction}
Our framework transforms unstructured text corpora into structured knowledge graphs to enable fine-grained knowledge evaluation. This transformation is crucial for capturing semantic relationships and facilitating precise knowledge auditing. Specifically, we construct two distinct knowledge graphs from the forget and retain datasets: \(\mathcal{G}_{\text{fgt}}\) and \(\mathcal{G}_{\text{ret}}\), respectively. Each knowledge graph represents a structured network of entities and their relationships, allowing for systematic analysis of knowledge units.
For implementation, following standard practices, we first segment the input text and perform coreference resolution preprocessing~\cite{lee2017end}, to ensure accurate entity identification and relationship mapping. We then employ the REBEL-large model~\cite{huguet-cabot-navigli-2021-rebel-relation}, which has been specifically fine-tuned for entity and relation extraction. This model demonstrates robust performance in extracting structured knowledge from natural language text, making it particularly suitable for our knowledge graph construction pipeline.

\subsection{Stage 2: Redundancy Removal}

The intricate entanglement of information across retain and forget datasets complicates the identification of specific elements requiring audit.
To address this challenge, we implement a graph alignment strategy to detect shared information between  \(\mathcal{G}_{\text{fgt}}\) and \(\mathcal{G}_{\text{ret}}\). We identify redundancy through triples that match exactly or share equivalent structures across both graphs. Our method examines each triple \((e_1, r, e_2) \in \mathcal{G}_{\text{fgt}}\) to locate its potential counterpart in \(\mathcal{G}_{\text{ret}}\). We express the overlapping edges mathematically as:
\begin{equation}
E_{\text{conf}} = E(\mathcal{G}_{\text{fgt}}) \cap E(\mathcal{G}_{\text{ret}}).
\end{equation}
The refined test graph is then constructed by removing these intersecting elements:
\begin{equation}
\mathcal{G}_{\text{test}} = \mathcal{G}_{\text{fgt}} \setminus E_{\text{conf}}.
\end{equation}
This process yields \(\mathcal{G}_{\text{test}}\), which maintains the fundamental structure of \(\mathcal{G}_{\text{fgt}}\) but excludes direct knowledge overlap with \(\mathcal{G}_{\text{ret}}\). The resulting graph provides a clean foundation for assessing selective forgetting performance, preserving crucial network relationships while eliminating redundant elements.
It is important to note that this step provides an approximation rather than a perfectly precise identification of redundant knowledge. Even if two facts appear to be identical, their meanings may vary depending on the surrounding context, making exact equivalence challenging to determine. Nevertheless, the distant supervision strategy employed here has been shown to effectively capture the majority of overlapping knowledge~\cite{mintz2009distant}. 




\subsection{Stage 3: Question Synthesis}
Previous benchmarks generate QA pairs by directly feeding entire text segments to LLMs, making it difficult to ensure comprehensive coverage and quality control of the resulting questions.  
To address this limitation, we adopt a fine-grained, dual-input prompting strategy. Specifically, for each knowledge triple in \(\mathcal{G}_{\text{test}}\), we leverage an LLM to automatically generate targeted test questions. 
Our dual-input prompting strategy equips LLMs with two complementary information sources: structured knowledge triples and their corresponding source text passages. This approach guides the model to generate fact-anchoring questions while maintaining fidelity to the original context. 
By anchoring question generation in both structured knowledge and source text, we ensure the generated questions accurately reflect the intended specific facts while preserving contextual relevance.
By enumerating each edge in \(\mathcal{G}_{\text{test}}\) and instructing the LLM to generate corresponding QA questions, we can guarantee at least a lower bound on the audit adequacy.

Our prompt design is based on several key principles. First, we explicitly define the LLM's role as an expert quiz question generator to set clear expectations. Second, by providing structured inputs consisting of both the knowledge triple and its original context, we ensure that the generated questions are firmly grounded in the relevant information. Third, we impose strict criteria on the generated questions: each must be answerable solely from the provided context, specific enough to yield a unique answer, and directly assess the semantic relationship between target entities. To facilitate automated evaluation, we require that each question-answer pair be output in a structured JSON format.

Furthermore, we adopt the one-shot learning by incorporating carefully selected example question-answer pairs into the prompt. These examples illustrate the desired question format and level of specificity, guiding the LLM toward generating high-quality, targeted questions. This comprehensive prompting strategy ensures that the synthesized questions effectively evaluate selective forgetting while maintaining human interpretability. The specific prompt employed in our experiments is provided in \autoref{sec:appendix_dataset}.

%% file: tfsrc/algo_method.tex
\begin{algorithm}[tb]
    \small
    \caption{\sys}
    \label{alg:selective_forgetting}
    \textbf{Input:} Forget dataset $D_{\text{fgt}}$, Retain dataset $D_{\text{ret}}$ \\
    \textbf{Output:} Audit suite $S$
    \begin{algorithmic}[1]
        \Function{GENERATION}{$D_{\text{fgt}}, D_{\text{ret}}$}
            \State $\triangleright$ Knowledge Graph Construction
            \State $G_{\text{fgt}} \gets \text{KGExtraction}(D_{\text{fgt}})$ 
            \State $G_{\text{ret}} \gets \text{KGExtraction}(D_{\text{ret}})$ 
            
            \State $\triangleright$ Redundancy Removal
            \State $G_{\text{test}} \gets \emptyset$
            \ForAll{$e \in G_{\text{fgt}}$}
                \If{$e \notin G_{\text{ret}}$}
                    \State $G_{\text{test}} \gets G_{\text{test}} \cup \{e\}$
                \EndIf
            \EndFor
            
            \State $\triangleright$ Question Synthesis
            \State $S \gets \emptyset$
            \ForAll{$e \in G_{\text{test}}$}
                \State $ctx \gets \text{RetrieveContext}(e)$ 
                \State $prompt \gets \text{ComposePrompt}(e, ctx)$ 
                \State $qa \gets \text{LLM}(prompt)$
                \State $S \gets S \cup \{qa\}$
            \EndFor
            
            \State \Return $S$
        \EndFunction
    \end{algorithmic}
    \end{algorithm}

%% file: src/exp.tex
\section{Experiments}
\label{sec:exp}

\subsection{Experimental Setup}
\label{sec:exp_setup}


Building upon MUSE, a comprehensive benchmark for LLM unlearning that provides extensive datasets and evaluation frameworks ~\cite{shi2024muse}, we integrate \sys to enhance its capabilities. For question generation, we leverage the state-of-the-art DeepSeek-V3 model~\cite{liu2024deepseek}, which has demonstrated superior performance in recent evaluations. The MUSE framework incorporates two primary data domains—NEWS and BOOKS—and includes a specially adapted LLaMA2-7B model that has undergone thorough training on the complete dataset. This fine-tuned model serves as the input for various unlearning techniques.

\textit{\textbf{Unlearning Methods.}} In our evaluation, we investigate three representative unlearning methods, each employing distinct strategies to achieve knowledge removal while preserving model utility.We utilize the default implementations, configurations, and scripts provided in MUSE~\cite{shi2024muse}.
Gradient Ascent(GA) operates by inverting the conventional training objective, maximizing the likelihood loss on forgotten data to discourage the generation of memorized content. 
Negative Preference Optimization (NPO) reframes the unlearning problem through the lens of preference optimization, treating forgotten knowledge as negative examples. 
Task Vectors (TV) implements unlearning through a novel weight arithmetic approach. The method first creates a reinforced model by training on forgotten content, then derives a task vector representing the direction of memorization. Unlearning is achieved by subtracting this vector from the original model weights, effectively steering the model away from the memorized information.
GA and NPO can be further enhanced with two utility preservation strategies: Gradient Descent on Retain set (GDR) and KL Divergence Regularization (KLR). 

\textit{\textbf{Metrics.}}
We evaluate the effectiveness of unlearning through our generated audit suite by quantifying the number of \textit{knowledge memorization cases (KMCs)} in the unlearned model. 
Unlike existing work that assess unlearning based on overall response similarity across the entire dataset, our method applies software testing principles to pinpoint specific failure-revealing test cases—scenarios in which an LLM provider might be liable for disclosing sensitive information. 
The identification process employs two complementary criteria for judgment. The first criteria uses ROUGE Recall to measure surface-level similarity, requiring model outputs to exceed a strict threshold (Recall=1) compared to reference answers. 
The second metric leverages an entailment-based approach ~\cite{yuan2024closer}, utilizing a pre-trained NLI model as described in ~\cite{sileo2024tasksource} to verify semantic equivalence between generated and reference answers without logical inconsistencies.
A higher frequency of detected memorization cases indicates less successful unlearning, while simultaneously demonstrating the comprehensiveness of our testing methodology.

\subsection{Details of Generated Audit Suite}

We applied \sys to two corpora provided by MUSE, namely the NEWS and BOOKS datasets.
The details are summarized in \autoref{tab:exp_datasetdetail}.
For the NEWS dataset, \sys extracted a knowledge graph (KG) from the forget dataset comprising 24,763 facts. After removing redundant knowledge, a final KG containing 16912 facts was obtained, from which 69,609 QA pairs were generated (On average, one fact corresponds to the generation of 4.11 QA pairs).
Similarly, for the BOOKS dataset, \sys extracted a KG with 41,123 facts from the forget dataset. Following the elimination of redundant knowledge, a final KG comprising 27,254 facts was produced, and 111,855 QA pairs were generated from this KG (on average, one fact corresponds to the generation of 4.10 QA pairs).
These results demonstrate the capability of \sys to automatically extract fine-grained knowledge graphs and generate large-scale audit suites.

\input{tfsrc/tab_datasetdetail.tex}
\input{tfsrc/tab_quality.tex}

\input{tfsrc/tab_failurenumber.tex}
\noindent\textbf{Mannual Assessment of the Generated Data.}
To rigorously assess the quality of HANKER's generated audit dataset, we conducted a detailed manual evaluation on randomly sampled 100 text chunks from each of the NEWS and BOOKS datasets. Our assessment focused on both the accuracy of extracted knowledge triples and the quality of generated QA pairs through four key metrics.
Accuracy of Knowledge Fact (AK) measures the precision of knowledge triple extraction from the source text, achieving scores of 0.76 and 0.61 for NEWS and BOOKS respectively. The relatively lower score on BOOKS reflects the inherent challenges in extracting structured knowledge from narrative text compared to more factual NEWS articles.
Question-Fact Relevance (QR) evaluates how well generated questions align with both the context and extracted facts. High scores of 0.91 (NEWS) and 0.84 (BOOKS) indicate that our framework effectively translates extracted knowledge into contextually appropriate questions.
Question Clarity (QC) assesses the linguistic quality and specificity of generated questions. Near-perfect scores of 0.99 across both domains demonstrate our system's exceptional ability to generate clear, unambiguous, and well-formed questions regardless of source material complexity.
Answer-Context Consistency (AC) gauges whether generated reference answers accurately reflect the source context. Strong performance of 0.91 (NEWS) and 0.84 (BOOKS) suggests reliable answer generation that maintains fidelity to the original text.
These results demonstrate \sys's capability in generating high-quality audit datasets, particularly excelling in question generation.

\input{tfsrc/fig_impact_redundancy.tex}

\subsection{Evaluation on Unlearning Methods}

Our result reveals a striking disparity in the ability to detect knowledge memorization cases between \sys's comprehensive audit suite and MUSE's baseline approach. The results paint a concerning picture about the extent of retained knowledge in supposedly unlearned models that were previously undetectable with limited audit sets.
On the NEWS dataset, \sys's detection capability proves remarkably more sensitive: using the ROUGE metric, it identifies over 4,600 memorization cases in the unmodified model, compared to just 33 cases detected by MUSE - a 142-fold increase in detection power. This gap widens even further when examining semantic understanding through the Entailment metric, where \sys detects more than 23,600 cases versus MUSE's 19 cases, representing a dramatic 1,242-fold improvement in identifying retained knowledge.
The BOOKS dataset tells an equally compelling story. \sys's comprehensive evaluation uncovers more than 4,700 memorization cases using ROUGE (compared to MUSE's 25 cases), and a remarkable 38,388 cases using Entailment (versus MUSE's 15 cases). These findings represent average improvements of 188× and 1,125× respectively in detection capability.

Particularly noteworthy is how these results persist across different unlearning methods. Even with state-of-the-art approaches like \(GA_{KLR}\) and \(NPO_{KLR}\), \sys consistently reveals significantly more cases where knowledge removal was incomplete.
This suggests that current unlearning methods may be less effective than previously thought, with their apparent success potentially being an artifact of insufficient testing rather than genuine knowledge removal.

These findings underscore the critical importance of comprehensive testing in evaluating unlearning effectiveness, revealing that the challenge of selective knowledge removal may be substantially more complex than indicated by previous benchmarks.

\subsection{Impact of Knowledge Redundancy on Unlearning Effectiveness Audits}
\label{sec:exp_conflict}

To validate the necessity of knowledge redundancy detection and elimination, we conducted a comprehensive experiment to assess its impact on unlearning evaluation effectiveness. Using the NEWS dataset as our testbed, we compared evaluation outcomes between two scenarios: one using the full dataset (126,224 test cases) and another using our deduplicated dataset (69,609 test cases). Our analysis considered both the number of identified knowledge memorization cases and standard dataset-level metrics (ROUGE and Entailment scores) used in existing evaluations.
The results reveal a striking impact of knowledge redundancy on evaluation outcomes. When using our deduplicated audit set, the number of identified knowledge memorization cases decreased substantially: detection rates dropped by 71.3-73.3\% under the ROUGE criterion and by 58.3-59.2\% under the Entailment criterion. This significant reduction suggests that knowledge redundancy leads to substantial false positives, where retained knowledge is incorrectly flagged as forgetting failures.
Furthermore, our analysis of quantitative metrics demonstrates that knowledge redundancy artificially inflates unlearning effectiveness measures. Without deduplication, ROUGE scores showed artificial inflation ranging from 19.7\% to 26.1\%, while Entailment scores were inflated by 32.4\% to 35.2\%. These inflated metrics indicate that traditional evaluation approaches may significantly overestimate unlearning effectiveness when redundant knowledge is not properly controlled for.

These findings provide compelling evidence for both the effectiveness of our approach and the critical importance of knowledge redundancy elimination in unlearning evaluation. The substantial reductions in false positives and metric inflation demonstrate that rigorous knowledge deduplication is essential for an accurate assessment of unlearning effectiveness.

%% file: tfsrc/tab_datasetdetail.tex
\begin{table}[tbp]
    \centering
    \small
    \setlength{\tabcolsep}{3pt}
    \caption{Statistics of Knowledge Extraction and QA Dataset}
    \begin{tabular}{ccccc}
    \toprule
    Dataset & \multicolumn{1}{c}{Initial Facts} & \multicolumn{1}{c}{Final Facts} & \multicolumn{1}{c}{QA Pairs} & \multicolumn{1}{c}{Average} \\
    \midrule
    News & 24,763 & 16,912 & 69,609 & 4.11 \\
    Books & 41,123 & 27,254 & 111,855 & 4.10 \\
    \bottomrule
    \end{tabular}
    \label{tab:exp_datasetdetail}
\end{table}

%% file: tfsrc/tab_quality.tex
\begin{table}[t]
    \centering
    \small 
    \caption{Quality assessment of generated knowledge graphs and QA pairs based on the following metrics: Knowledge Fact Accuracy (AK), Question–Fact Relevance (QR), Question Clarity (QC), and Answer–Context Consistency (AC).}
    \label{tab:quality_qa}
    \begin{tabular}{ccccc}
    \hline
          & AK  & QR   & QC   & AC   \\ \hline
    NEWS  & 0.76 & 0.91 & 0.99 & 0.91 \\
    BOOKS & 0.61 & 0.84 & 0.99 & 0.84 \\ \hline
    \end{tabular}
    \end{table}

%% file: tfsrc/tab_failurenumber.tex
\begin{table}[!t]
    \small
    \caption{Numbers of Knowledge Memorization Cases on News.}
    \label{tab:KMC_news}
\begin{tabular}{lllllllll}
\cline{1-5}
\multirow{2}{*}{Method} & \multicolumn{2}{c}{\textbf{MUSE}} & \multicolumn{2}{c}{\textbf{\sys}} &  &  &  &  \\ \cline{2-5}
                    & ROUGE      & Entail.     & ROUGE             & Entail.             &  &  &  &  \\ \cline{1-5}
w/o unlearn  & 33         & 19          & 4688              & 23605               &  &  &  &  \\
\(GA_{KLR}\)  & 18         & 3          & 3702              & 21650               &  &  &  &  \\
\(NPO_{GDR}\) & 27         & 13          & 4454              & 23474               &  &  &  &  \\
\(NPO_{KLR}\) & 19         & 6        & 3780              & 21571               &  &  &  &  \\
Task Vector  & 33         & 10          & 4853              & 23808               &  &  &  &  \\ \cline{1-5}
\end{tabular}
\end{table}

\begin{table}[!t]
    \small
    \caption{Numbers of Knowledge Memorization Cases on Books.}
    \label{tab:KMC_books}
\begin{tabular}{lllllllll}
\cline{1-5}
\multirow{2}{*}{Method} & \multicolumn{2}{c}{\textbf{MUSE}} & \multicolumn{2}{c}{\textbf{\sys}} &  &  &  &  \\ \cline{2-5}
                        & ROUGE      & Entail.     & ROUGE             & Entail.             &  &  &  &  \\ \cline{1-5}
    w/o unlearn  & 25         & 15          & 4729              & 38388              &  &  &  &  \\
    \(GA_{KLR}\)  & 6          & 7          & 3490              & 32365               &  &  &  &  \\
    \(NPO_{GDR}\) & 0          & 34          & 1435              & 18094               &  &  &  &  \\
    \(NPO_{KLR}\) & 4          & 8          & 3447              & 32332               &  &  &  &  \\
    Task Vector  & 25         & 15          & 4700              & 38210               &  &  &  &  \\ \cline{1-5}
    \end{tabular}
\end{table}

%% file: tfsrc/fig_impact_redundancy.tex
\begin{figure*}[t!]
    \centering
    \begin{minipage}{0.49\textwidth}
        \centering
        \includegraphics[width=\linewidth]{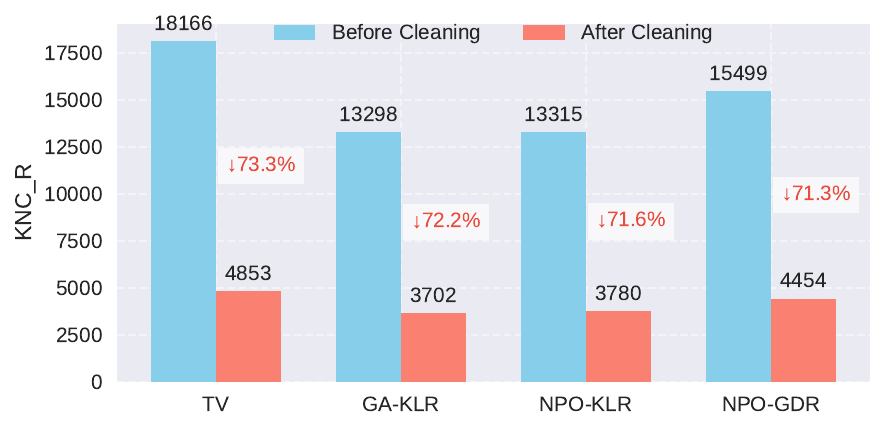}
        \caption*{(a) Number of KMCs (by Rouge)}
    \end{minipage}
    \hfill
    \begin{minipage}{0.49\textwidth}
        \centering
        \includegraphics[width=\linewidth]{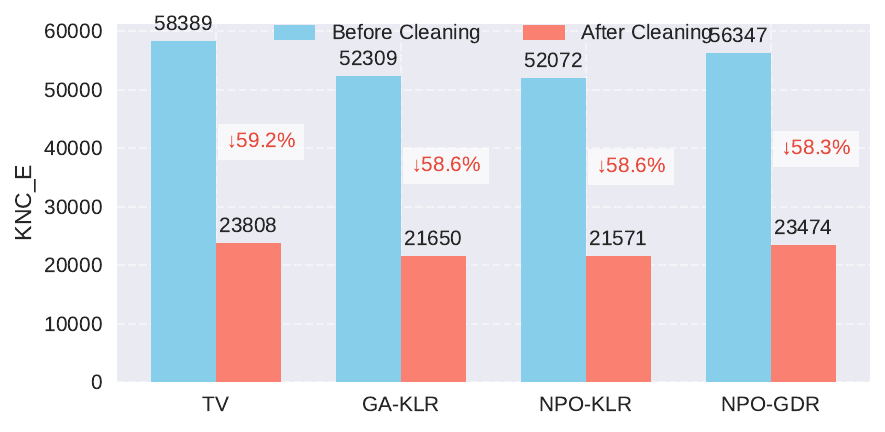}
        \caption*{(b) Number of KMCs (by Entailment)}
    \end{minipage}
    
    \vspace{5mm}

    \begin{minipage}{0.49\textwidth}
        \centering
        \includegraphics[width=\linewidth]{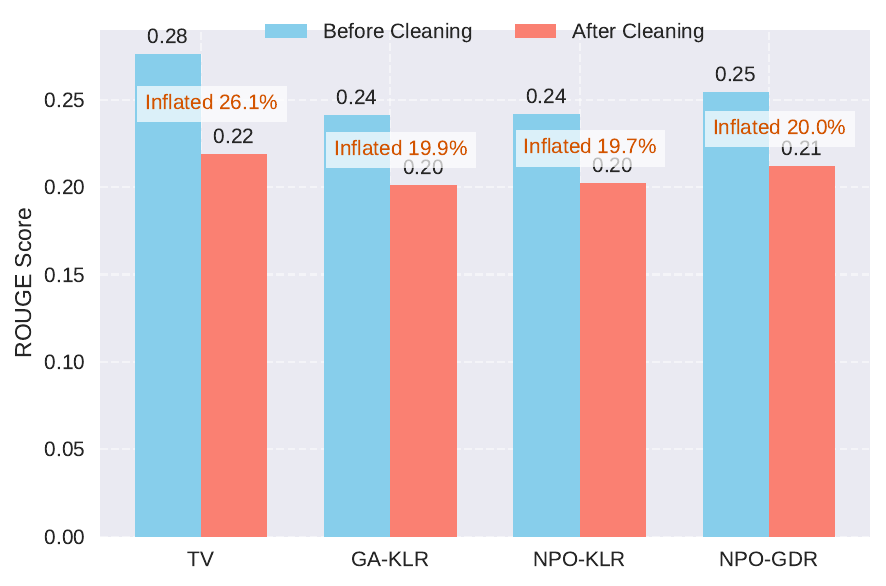}
        \caption*{(c) ROUGE Score}
    \end{minipage}
    \hfill
    \begin{minipage}{0.49\textwidth}
        \centering
        \includegraphics[width=\linewidth]{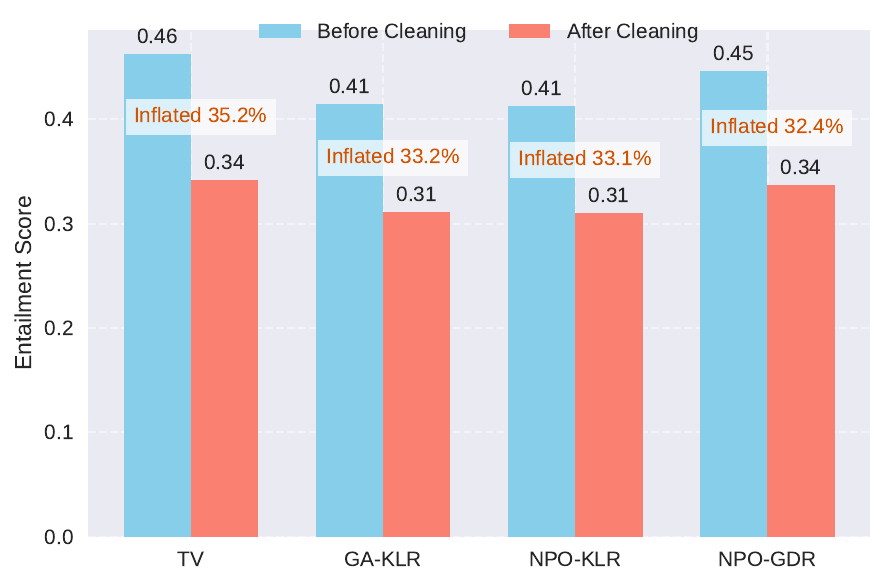}
        \caption*{(d) Entailment Score}
    \end{minipage}
    
    \caption{Impact of Redundancy on Knowledge Memorization Cases.}
    \label{fig:impact_redundancy}
\end{figure*}

%% file: src/rw.tex
\section{Related Work}
\label{sec:rw}

\noindent\textbf{Machine Unlearning for LLMs}. Machine unlearning, a technique first established for classification challenges~\cite{bourtoule2021machine}, has progressively evolved toward applications in large language models. Contemporary research predominantly explores parameter optimization methodologies, achieved through targeted fine-tuning procedures~\cite{yao2023large, jang2022knowledge,wang2024selective,yao2024machine,tian2024forget, liu2024learning,gu2024second, jia2024soul}
The transparent nature of modifying neural architectures engenders enhanced user trust, despite potential compromises to overall model performance. Beyond parameter-based approaches, researchers have pioneered diverse methodologies including advanced contrastive decoding frameworks~\cite{eldan2023s,wang2024rkld,ji2024reversing,huang2024offsetunlearninglargelanguage}, 
task-specific vector implementations~\cite{liu2024saferlargelanguagemodels,dou2025avoidingcopyrightinfringementlarge}, contextual learning strategies~\cite{pawelczyk2024incontextunlearninglanguagemodels,muresanu2024unlearnablealgorithmsincontextlearning}, and sophisticated input processing mechanisms~\cite{gao2024practicalunlearninglargelanguage, liu2024largelanguagemodelunlearning}. 

\noindent\textbf{Evaluation of LLM Unlearning.}
The evaluation unlearning effectiveness of LLM encompasses diverse task scenarios. Early research focused on traditional NLP classification tasks to examine models' prediction~\cite{chen2023unlearnwantforgetefficient}. Subsequently, researchers developed specialized datasets to provide standardized evaluation platforms~\cite{eldan2023s, shi2024muse,maini2024tofu}. 
Besides some work has been devoted to focusing on the robustness of unlearning, i.e., adding perturbations or rewrites to the same problem to activate model memory~\cite{joshi-etal-2024-towards}.

\noindent\textbf{Knowledge Graphs for Evaluation.}
Knowledge graphs offer distinct advantages beyond the completeness and identifiability properties utilized in this study. They serve as effective tools for evaluating both QA systems~\cite{Wangkgdi2024} and LLM unlearning~\cite{wu2024evaluatingdeepunlearninglarge}. Notably, knowledge graphs enable the assessment of model reasoning capabilities through transitive relationships (if a→b and b→c, then testing whether the model infers a→c).
The framework we propose in this paper conveniently integrates with these techniques.

%% file: src/conclusion.tex
\section{Conclusion}
\label{sec:conclusion}

In this paper, we introduce \sys, an automated framework for generating holistic audit datasets to evaluate the effectiveness of LLM unlearning. By leveraging knowledge graphs, HANKER addresses two critical challenges in unlearning evaluation: ensuring audit adequacy and eliminating knowledge redundancy between the forget and retain datasets. 
Our empirical analysis on the popular MUSE benchmark demonstrates that \sys can significantly expand the scale of audit datasets, identifying thousands of knowledge memorization cases that previous benchmarks failed to detect, and revealing how
knowledge redundancy significantly skews unlearning effectiveness metrics.

%% file: src/limitation.tex
\section*{Limitations and Ethical Considerations}

\noindent\textbf{Limitations.} The primary limitation of our work is that it extends only the dataset provided by MUSE and employs DeepSeek-v3 for question generation. 
To mitigate this generalization risk, we have released our code and the generated audit suite, allowing researchers to utilize our framework to create additional audit datasets and evaluate their quality. Meanwhile, this is also our future work to extend our framework to other benchmarks.

\noindent\textbf{Ethical Considerations.} Machine unlearning can be employed to mitigate risks associated with LLMs in terms of privacy, security, bias, and copyright. Our work is dedicated to providing a comprehensive evaluation framework to help researchers better understand the unlearning effectiveness of LLMs, which we believe will have a positive impact on society.

%% file: src/appendix.tex
\section{Appendix}
\label{sec:appendix}

\subsection{Dataset Details}
\label{sec:appendix_dataset}

Below, we present the specific prompts used with DeepSeek-V3 for generating audit questions.
\input{tfsrc/code_prompt.tex}

%% file: tfsrc/code_prompt.tex
\begin{figure*}[htbp]
\centering
\begin{lstlisting}[language=Python]
SYS_PROMPT = """You are an expert quiz generator. Given a text passage and a relationship triple, generate specific questions to test knowledge about this relationship based on the context provided.

Input Format:
- Text: A passage containing information about the relationship
- Relationship: A triple containing {'head': entity1, 'type': relation_type, 'tail': entity2}

Task:
Generate up to 5 focused questions that test understanding of the relationship between the head entity and tail entity, considering:
1. Questions should be answerable solely from the given context
2. Questions should be specific enough to have a unique correct answer
3. Questions can ask about the tail entity given the head entity and relationship type
4. Questions can ask about the relationship between the two entities
5. Questions can ask about specific details that establish this relationship

Requirements:
1. Each question must have a clear, unambiguous answer based on the context
2. Avoid overly broad or general questions
3. Focus on the specific relationship provided
4. Use the context to add specific details to questions
5. Ensure questions and answers are factually consistent with the provided text

Response Format:
The response must be a valid JSON object with the following structure:
{
    "1": {
        "question": "Your question text here",
        "reference_answer": "The correct answer based on context"
    },
    "2": {
        "question": "...",
        "reference_answer": "..."
    }
    // ... up to 5 questions
}

Example Input:
Text: "The Greek Orthodox Church observes Lent as a period of fasting and spiritual reflection that begins on Clean Monday and lasts for 40 days. During this time, adherents follow strict dietary restrictions and increase their prayer and attendance at special services."
Relationship: {'head': 'Lent', 'type': 'religion', 'tail': 'Greek Orthodox'}

Example Output:
{
    "1": {
        "question": "Which religious denomination observes Lent beginning on Clean Monday with a 40-day period of fasting and spiritual reflection?",
        "reference_answer": "Greek Orthodox"
    },
    "2": {
        "question": "In the Greek Orthodox tradition, what is the length of the Lent period?",
        "reference_answer": "40 days"
    }
}
"""

USER_PROMPT = """
Please generate questions based on the following input:

Text: {text}
Relationship: {relationship}
"""
\end{lstlisting}
\caption{Our prompt.}
\end{figure*}